\documentclass[11pt]{article}

\usepackage[margin=1in]{geometry}
\usepackage{amsmath}
\usepackage{amssymb}
\usepackage{booktabs}
\usepackage{graphicx}
\usepackage{url}
\usepackage{microtype}
\usepackage{natbib}
\usepackage{hyperref}

\title{\textbf{Evidence-State Reliability Under Controlled Degradation:\\
Parser-Validity Divergence in a Multi-Stage LLM Pipeline}}

\author{Naimur Rahman}

\date{August 2026}
\usepackage[margin=1in]{geometry}
\usepackage{amsmath}
\usepackage{amssymb}
\usepackage{booktabs}
\usepackage{graphicx}
\usepackage{url}
\usepackage{microtype}
\usepackage{natbib}
\usepackage{float}
\usepackage{hyperref}
\hypersetup{hidelinks}
\begin{document}

\maketitle

\begin{abstract}
Multi-stage LLM pipelines can remain structurally valid even when evidence available to downstream stages becomes incomplete, compressed, or conflicting. This paper introduces and operationalises Evidence-State Reliability (ESR), an evaluation layer concerned with whether intermediate evidence remains sufficiently complete, grounded, internally consistent, and usable for a stage's assigned function. ESR is evaluated separately from parser validity, which measures structural conformance.

We evaluate the framework using GLM-5.2 on 60 sanitised base cases under four evidence conditions: clean, compressed-lossy, partial-dropout, and noisy-conflicting. Each condition was processed through decision, audit, and escalation stages. The design comprised 720 planned and ledgered calls, with 713 retained sanitised execution rows.

Across nine matched degraded-minus-clean condition-stage comparisons, all operational stage-success estimates were negative and all 95\% bootstrap intervals remained below zero. All nine parser-validity point estimates were positive, although the three partial-dropout intervals included zero. Among parser-valid degraded audit outputs, degradation detection was 1.0 in each degraded condition while false-assurance rates remained non-zero; among parser-valid degraded escalation outputs, recovery was 0.0 in every degraded condition.

The results show a bounded reliability-layer divergence in the evaluated pipeline: structural conformance can improve directionally while evidence-sensitive stage success deteriorates under the same controlled intervention. They also separate detection of degraded evidence from recovery. The conclusions are limited to the evaluated model configuration, pipeline design, selected sanitized cases, scoring procedure, and single scaled run.
\end{abstract}


\section{Introduction}
\label{sec:introduction}

Large language models are increasingly used as components of multi-stage systems rather than only as isolated question-answering tools. A pipeline may receive or retrieve evidence, transform that evidence into an intermediate representation, generate a decision, inspect the decision through an audit component, and route difficult cases to an escalation mechanism. In such systems, reliability depends not only on the final output but also on the evidence transferred between stages.

Evidence can deteriorate before it reaches a downstream component. Relevant information may be removed during compression, omitted through partial data loss, or made difficult to reconcile by conflicting signals. A later stage may therefore receive an evidence state that is no longer adequate for its assigned function while continuing to produce fluent, syntactically valid, and machine-readable output.

This creates a reliability problem that structural validation alone cannot capture. A parser-valid response satisfies a formal output contract, such as syntactically valid JSON containing the required fields. Parser validity does not establish that the available evidence was sufficient, appropriately grounded, internally consistent, or usable for the substantive function assigned to the stage. A structurally valid object can therefore continue through an automated pipeline without producing a technical exception even when its evidential basis has deteriorated.

This paper introduces and operationalises \emph{Evidence-State Reliability} (ESR) as an evaluation layer for multi-stage AI pipelines. ESR concerns whether the evidence supplied to a stage remains sufficiently complete, grounded, internally consistent, and usable for the evidence-sensitive function assigned to that stage. The intermediate evidence state is therefore treated as an explicit object of evaluation rather than being assessed only indirectly through the final output.

ESR is not intended to replace accuracy, calibration, robustness, faithfulness, parser validation, or explanation assessment. These measures evaluate different properties of system behaviour. Parser validity asks whether an output satisfies a structural contract. ESR concerns whether the evidence available to a stage can support the function assigned to that stage. In the present study, stage-specific success is used as an operational indicator of ESR, but it is not treated as an exhaustive measurement of the broader construct.

The central research question is:

\begin{quote}
\emph{How can Evidence-State Reliability be measured separately from parser validity in a multi-stage LLM decision pipeline?}
\end{quote}

Three supporting research questions guide the analysis:

\begin{enumerate}
    \item[\textbf{RQ1.}] Do parser validity and evidence-sensitive stage success respond differently to controlled evidence degradation?

    \item[\textbf{RQ2.}] Does successful detection of degraded evidence at an audit stage lead to recovery at a subsequent escalation stage?

    \item[\textbf{RQ3.}] What sequence-level failure patterns become visible when decision, audit, and escalation outcomes are analysed together rather than as isolated outputs?
\end{enumerate}

To investigate these questions, the study uses a controlled within-case experiment with 60 sanitised base cases derived from the context of the Consumer Financial Protection Bureau Consumer Complaint Database. Each base case was represented under four evidence conditions: a clean experimental baseline, compressed-lossy evidence, partial-dropout evidence, and noisy-conflicting evidence. Each condition was processed through three pipeline stages: decision, computational audit, and escalation. The planned experimental design therefore comprised

\[
60 \text{ cases}
\times 4 \text{ evidence conditions}
\times 3 \text{ stages}
= 720 \text{ model calls}.
\]

Base-case identity was preserved across conditions so that each degraded observation could be compared with the corresponding clean observation for the same case and pipeline stage.

The experiment examines whether structural conformance and evidence-sensitive stage success respond differently to the same evidence interventions. Across the nine matched degraded-minus-clean condition-stage comparisons, all operational stage-success estimates were negative and all corresponding 95\% bootstrap intervals remained below zero. In contrast, all nine parser-validity point estimates were positive, although the three partial-dropout parser-validity intervals included zero. Within the evaluated pipeline, structural conformance therefore moved in a favourable direction while evidence-sensitive stage success deteriorated under the same controlled interventions.

The downstream stages revealed a second distinction. Among parser-valid degraded audit outputs, the implemented audit indicator detected degradation in every degraded condition. Detection, however, did not imply recovery. Audit false-assurance rates remained non-zero, and no parser-valid degraded escalation output was classified as recovered. Detection of an evidential problem and restoration of a successful evidence-sensitive outcome are therefore treated as separate pipeline functions in this study.

The study makes four bounded contributions. First, it operationalises parser validity, evidence-sensitive stage success, reliability-layer divergence, and format-layer false assurance as distinct measurements within a stage-aware evaluation framework. Second, it applies controlled evidence-state interventions across a decision--audit--escalation sequence while preserving base-case identity across experimental conditions. Third, it reports matched condition-stage comparisons and nonparametric bootstrap intervals rather than relying only on aggregate performance values. Fourth, it connects stage-level behaviour to sequence-level reliability patterns that distinguish parser failure, detected-but-unrecovered degradation, audit false assurance, incomplete execution, and preserved success.

The resulting framework is intended as a diagnostic approach to evaluating intermediate evidence states and stage transitions. The findings do not establish universal LLM unreliability or show that evidence degradation generally improves parser validity. They remain confined to the evaluated GLM-5.2 configuration, pipeline design, controlled evidence conditions, selected sanitised cases, scoring procedure, and single scaled experimental run. The study likewise does not establish deployment safety, regulatory validity, complaint truth, consumer-harm prevalence, provider-independent behaviour, or cross-model generality.

\section{Related Work}
\label{sec:related-work}

\subsection{Evaluation beyond aggregate correctness}

Reliability cannot always be represented by a single aggregate performance measure. Behavioural testing frameworks such as CheckList decompose model performance into capability-oriented evaluations rather than assuming that held-out accuracy captures every relevant aspect of behaviour \citep{ribeiro-etal-2020-beyond}. Calibration research similarly distinguishes predictive correctness from the reliability of reported confidence estimates \citep{pmlr-v70-guo17a}.

Retrieval-augmented generation evaluation follows a related multidimensional approach. RAGAS separates properties including faithfulness, context relevance, and answer relevance, while ARES evaluates multiple components contributing to retrieval-augmented generation performance \citep{es-etal-2024-ragas,saad-falcon-etal-2024-ares}. These approaches illustrate a broader principle relevant to the present study: an output may satisfy one evaluation criterion while remaining deficient under another.

Evidence-State Reliability adopts this layered perspective but focuses on the evidence available at intermediate stages of a pipeline. Rather than treating final-answer quality as the only object of evaluation, ESR examines whether a stage receives evidence capable of supporting its assigned evidence-sensitive function.

\subsection{Structural validity and evidence sufficiency}

Structured outputs are important when language-model responses must be consumed automatically by downstream software. Grammar-constrained decoding and related approaches can improve adherence to formal output specifications \citep{geng-etal-2023-grammar}. Work on structured-output evaluation further distinguishes formal validity from properties such as value correctness, executability, and task performance \citep{ray2026constraint,singh2026structured}.

The present study does not claim to originate the distinction between structural validity and substantive adequacy. Instead, it examines how structural validity and evidence-sensitive stage success respond to controlled changes in the evidence supplied to the same pipeline. Parser validity is measured independently at the decision, audit, and escalation stages, allowing its response to evidence degradation to be compared with evidence-sensitive success under the same condition-stage combinations.

Evidence sufficiency has also been studied in fact checking, retrieval, and selective generation. Work on fact checking with insufficient evidence demonstrates the importance of distinguishing claims that can be supported from those for which the available evidence is inadequate \citep{atanasova-etal-2022-fact}. Related work on sufficiency-aware retrieval and delayed ground truth further considers how evidence availability can affect whether a task is answerable or can be acted upon reliably \citep{qiu2026surerag,solozobov2026evidence}.

ESR extends this concern from the evidence supporting an endpoint answer to evidence states within a multi-stage sequence. In the present experiment, controlled versions of the same base case are processed through decision, audit, and escalation stages, making it possible to examine not only whether degradation affects stage success, but also whether it is detected and whether a downstream mechanism recovers from it.

\subsection{Component-aware evaluation, auditing, and recovery}

Evaluation of multi-stage and agentic systems increasingly considers component-level behaviour rather than only final outputs. AgentBench evaluates capabilities required by interactive language-model agents \citep{liu2024agentbench}, while retrieval-augmented generation frameworks separately assess components contributing to an eventual response \citep{es-etal-2024-ragas,saad-falcon-etal-2024-ares}. Component-level evaluation can reveal where a failure appears in a sequence and how later stages respond to it.

The term \emph{audit} requires particular care. Algorithmic-audit research addresses broader organisational processes involving documentation, accountability, governance, and institutional practice \citep{raji2020closing}. The audit stage in this study is narrower: it is a computational, model-output-coded component that records whether evidence degradation was detected and whether an experimental false-assurance indicator appeared. It is not an independent institutional audit and does not establish regulatory assurance.

Selective prediction and learning-to-defer research similarly distinguish prediction from abstention, referral, and expert intervention \citep{pmlr-v97-geifman19a,pmlr-v119-mozannar20b}. The escalation stage evaluated here is not a learned optimal deferral policy. Its role is to measure whether the implemented downstream mechanism recovered after degradation had entered the pipeline. This distinction is central to the present design because detecting a reliability problem and restoring successful operation are separate outcomes.

\subsection{Reliability cascades}

Cascading failure is a long-standing systems concept in which local disruptions propagate through connected structures \citep{motter2002cascade}. Recent work on multi-agent language-model systems has likewise examined how errors or hallucinations can propagate between interacting components or agents \citep{xie2026spark,jamshidi2026hallucination}.

The term \emph{reliability cascade} is used more narrowly in this study. It refers to an operational sequence-level pattern in which parser failure, evidence-sensitive failure, detected-but-unrecovered degradation, false assurance, or incomplete execution prevents a preserved successful path through the evaluated decision--audit--escalation sequence.

This taxonomy is diagnostic rather than a general theory of cascading failure. It distinguishes structurally different sequence outcomes so that parser failure, detected-but-unrecovered degradation, false assurance, and incomplete execution are not collapsed into a single endpoint score. The resulting cascade rate is therefore interpreted only as a summary of the evaluated experimental sequences, not as an estimate of failure prevalence in deployed AI systems.

\subsection{Positioning of this work}

The literature above addresses several components relevant to Evidence-State Reliability: multidimensional evaluation, structured-output validity, evidence sufficiency, component-level assessment, computational auditing, selective prediction, deferral, recovery, and cascading failure. The present study does not claim that these individual ideas are new.

Its contribution is their bounded integration within a single controlled and matched multi-stage experiment. The framework jointly evaluates controlled evidence-state degradation, parser validity, evidence-sensitive stage success, reliability-layer divergence, audit detection, false assurance, escalation recovery, paired uncertainty, and sequence-level failure patterns while preserving base-case identity across evidence conditions.

The resulting contribution is therefore methodological and empirical rather than a claim of global priority. This work operationalises these distinctions as a stage-aware evaluation framework for examining how structural and evidence-sensitive reliability measures respond to the same controlled evidence interventions.

\section{Evidence-State Reliability Framework}
\label{sec:esr-framework}

\subsection{Evidence states and stage-specific objectives}

Let \(E_{i,c,k}\) denote the evidence state associated with base case \(i\), evidence condition \(c\), and pipeline stage \(k\).

An evidence state may contain original records, extracted fields, summaries, retrieved passages, intermediate conclusions, structured observations, or transformed representations. Its reliability cannot be assessed independently of the objective assigned to the stage receiving it. An evidence state that is adequate for a broad classification may, for example, be insufficient for a detailed audit or recovery action.

In this study, Evidence-State Reliability concerns whether the evidence available to a stage remains:

\begin{itemize}
    \item sufficiently complete for the assigned objective;
    \item grounded in the available source material;
    \item internally consistent enough to support interpretation; and
    \item usable by the receiving stage for its intended function.
\end{itemize}

These properties define the broader ESR construct. They are not assumed to be exhaustively represented by any single experimental variable. The primary operational indicator used in this experiment is stage-specific success under the committed scoring contract.

\subsection{Parser validity}

Let

\[
V_{i,c,k} \in \{0,1\}
\]

indicate whether the output associated with base case \(i\), condition \(c\), and stage \(k\) is parser-valid.

Parser validity means that the sanitised model output satisfies the required structural contract. A parser-valid output can, for example, contain syntactically valid JSON from which all required fields can be successfully parsed.

Let \(I_{c,k}\) denote the retained observations for condition \(c\) and stage \(k\), with

\[
N_{c,k} = |I_{c,k}|.
\]

The parser-validity rate is

\[
\widehat{\mathrm{PV}}_{c,k}
=
\frac{1}{N_{c,k}}
\sum_{i \in I_{c,k}}
V_{i,c,k}.
\]

Parser validity is necessary for dependable automated consumption, but it does not measure evidence sufficiency, substantive correctness, complaint truth, regulatory validity, or real-world decision quality.

\subsection{Operational ESR rate}

Let

\[
R_{i,c,k} \in \{0,1\}
\]

denote evidence-sensitive stage success under the committed scoring contract.

A retained observation is classified as stage-successful when both of the following conditions are satisfied:

\begin{enumerate}
    \item the output is parser-valid; and
    \item the sanitised \texttt{validity\_judgment} field is positive.
\end{enumerate}

Rows without a parser-valid output or a positive validity judgment are not counted as successful.

The operational ESR rate for condition \(c\) and stage \(k\) is

\[
\widehat{\mathrm{ESR}}_{c,k}
=
\frac{1}{N_{c,k}}
\sum_{i \in I_{c,k}}
R_{i,c,k}.
\]

The notation \(\widehat{\mathrm{ESR}}\) refers specifically to the operational indicator used in this experiment. It does not imply that stage success exhaustively measures every dimension of Evidence-State Reliability.

The analysis also retains a separate binary field,
\texttt{evidence\_state\_adequate}. Let

\[
A_{i,c,k} \in \{0,1\}
\]

denote this model-output-coded indicator. \(A_{i,c,k}\) records whether the supplied evidence state was coded as adequate for the relevant stage, whereas \(R_{i,c,k}\) records whether the complete stage-success criterion was satisfied.

The evidence-adequacy indicator is therefore distinct from operational ESR. It is also not an independent human adjudication of evidence quality. Matched degraded-minus-clean changes in evidence-state adequacy are calculated using the same paired procedure applied to parser validity and operational ESR.

\subsection{Matched changes under evidence degradation}

Let \(c_0\) denote the clean evidence condition and \(c\) a degraded condition. For each base case available under both conditions at stage \(k\), define the case-level change in parser validity as

\[
\delta^{\mathrm{PV}}_{i,c,k}
=
V_{i,c,k}
-
V_{i,c_0,k}.
\]

Let \(P_{c,k}\) denote the set of base cases for which both the clean and degraded observations are retained at stage \(k\), and let

\[
n_{c,k} = |P_{c,k}|.
\]

The matched mean change in parser validity is

\[
\Delta \mathrm{PV}_{c,k}
=
\frac{1}{n_{c,k}}
\sum_{i \in P_{c,k}}
\delta^{\mathrm{PV}}_{i,c,k}.
\]

The corresponding case-level change in operational ESR is

\[
\delta^{\mathrm{ESR}}_{i,c,k}
=
R_{i,c,k}
-
R_{i,c_0,k},
\]

with matched mean

\[
\Delta \mathrm{ESR}_{c,k}
=
\frac{1}{n_{c,k}}
\sum_{i \in P_{c,k}}
\delta^{\mathrm{ESR}}_{i,c,k}.
\]

This formulation preserves base-case identity: each degraded observation is compared with the corresponding clean observation for the same case and pipeline stage.

\subsection{Reliability-layer divergence}

Reliability-layer divergence (RLD) measures the difference between the response of structural validity and the response of evidence-sensitive stage success to the same evidence intervention.

For degraded condition \(c\) at stage \(k\),

\[
\mathrm{RLD}_{c,k}
=
\Delta \mathrm{PV}_{c,k}
-
\Delta \mathrm{ESR}_{c,k}.
\]

A positive RLD value indicates that parser validity changed more favourably than operational ESR.

The pattern of primary interest in this study is

\[
\Delta \mathrm{PV}_{c,k} > 0
\qquad \text{and} \qquad
\Delta \mathrm{ESR}_{c,k} < 0.
\]

Under this pattern, structural conformance improves directionally while evidence-sensitive stage success deteriorates under the same intervention. The two evaluation layers therefore provide opposing assessments of the intervention.

RLD does not replace either component measure. Its purpose is to make divergence between structural and evidence-sensitive behaviour explicit.

\subsection{False assurance at the format layer}

False assurance at the format layer occurs when an output is parser-valid but does not satisfy the operational evidence-sensitive success criterion.

The general format-layer false-assurance rate is

\[
\widehat{\mathrm{FA}}_{c,k}
=
\frac{1}{N_{c,k}}
\sum_{i \in I_{c,k}}
V_{i,c,k}
\left(1-R_{i,c,k}\right).
\]

This measure identifies outputs that satisfy the structural contract while remaining unsuccessful under the committed stage-specific scoring rule.

The audit-specific false-assurance result reported later in the paper uses a different calculation. Its denominator consists only of parser-valid audit rows, and its numerator consists of those rows for which the committed
\texttt{audit\_false\_assurance} field is positive. The audit-specific measure is therefore a model-output-coded pipeline indicator and should not be interpreted as independent adjudication that an output was incorrect or misleading.

\subsection{Audit detection and escalation recovery}

Audit detection is defined among parser-valid audit outputs. It records the proportion for which

\[
\texttt{audit\_detected\_degradation} = \texttt{true}.
\]

Escalation recovery is defined among parser-valid escalation outputs. It records the proportion for which

\[
\texttt{escalation\_recovery} = \texttt{true}.
\]

These variables represent different pipeline functions. Audit detection records whether the computational audit stage identified evidence degradation. Escalation recovery records whether the subsequent stage restored a successful evidence-sensitive outcome.

A detected problem is therefore not treated as equivalent to a recovered problem. Keeping the two measurements separate allows the analysis to identify cases in which degradation is recognised while the downstream mechanism remains unable to restore successful operation.

\subsection{Sequence-level reliability cascades}

A sequence is constructed for each base-case-by-condition combination using the corresponding decision, audit, and escalation outcomes.

The operational cascade taxonomy assigns sequences to five mutually exclusive outcome families:

\begin{enumerate}
    \item \textbf{Parser-failure cascade:} one or more required stage outputs fail the structural contract.

    \item \textbf{Detected-but-unrecovered degradation:} degradation is detected, but the escalation stage does not restore operational success.

    \item \textbf{Audit false assurance:} the sequence contains a positive audit false-assurance indicator under the committed scoring contract.

    \item \textbf{Incomplete persisted sequence:} one or more expected retained stage rows are missing.

    \item \textbf{Preserved success:} the sequence completes without meeting any of the operational cascade-failure criteria.
\end{enumerate}

The taxonomy is diagnostic rather than causal. It classifies the observed sequence pattern but does not independently establish the mechanism that produced it.

The categories distinguish different reliability problems. Parser failure concerns structural output or validation. Detected-but-unrecovered degradation indicates that recognition was not accompanied by successful remediation. Audit false assurance concerns a nominally valid audit output carrying the committed false-assurance indicator. Incomplete sequences represent execution or persistence failures.

Separating these outcomes prevents structurally different reliability problems from being collapsed into a single end-to-end failure measure.

\section{Methodology}
\label{sec:methodology}

\subsection{Study design}

The primary study used a paired, controlled design centred on 60 sanitised base cases. Each base case was represented under four evidence conditions and processed through three pipeline stages. The planned execution therefore comprised

\[
60 \text{ cases}
\times
4 \text{ evidence conditions}
\times
3 \text{ pipeline stages}
=
720 \text{ model calls}.
\]

Preserving base-case identity across conditions allowed each degraded observation to be compared with the corresponding clean observation for the same case and pipeline stage.

The 60 cases were randomly selected from a pre-specified, version-controlled sanitised evidence-state package. Every selected case contained all four recognised evidence conditions. The execution manifest therefore contained 240 condition-linked evidence states:

\begin{itemize}
    \item 60 clean;
    \item 60 compressed-lossy;
    \item 60 partial-dropout; and
    \item 60 noisy-conflicting.
\end{itemize}

The selected cases are not treated as a representative or probability sample of consumer complaints, financial decisions, or deployed AI-system interactions. The random selection was performed within the pre-specified experimental package rather than from a defined population of real-world consumer complaints. The cases therefore form a controlled experimental set for evaluating how the implemented pipeline responds to defined evidence-state interventions.

\subsection{Framework development and primary-study boundary}

The ESR framework was refined through preliminary simulation, real-model, sensitivity, and deterministic-domain studies. These developmental studies informed the evidence-condition definitions, the separation of decision, audit, and escalation functions, and the operational reliability-cascade taxonomy.

The developmental programme included a 750-row simulation study, a 60-chain GLM-5.2 pilot, a smaller Claude Opus 4.8 comparison subset, and a 72-chain deterministic second-domain study with robustness checks. These studies were exploratory and developmental. Their observations are not pooled with the primary experiment, treated as independent replications, or used to increase the effective sample size reported in this paper.

The principal empirical evidence analysed here is therefore the scaled 720-call GLM-5.2 experiment.

\subsection{Evidence provenance and sanitisation}

The evidence packets were derived from a local export of the Consumer Financial Protection Bureau Consumer Complaint Database downloaded on 7 July 2026 \citep{cfpb2026complaints}.

Case identifiers and evidence materials were sanitised before experimental execution and reporting. The shared reporting boundary excluded:

\begin{itemize}
    \item raw CFPB records;
    \item raw prompts;
    \item raw model responses;
    \item JSONL output archives;
    \item API credentials; and
    \item environment files.
\end{itemize}

CFPB provenance was used to provide a substantively realistic complaint-context substrate. It was not used to establish whether an individual complaint was factually correct, whether a company acted unlawfully, whether a regulatory violation occurred, or whether an actual financial decision was appropriate.

The experimental units are sanitised evidence states and model-output-coded stage outcomes. The study therefore evaluates a controlled pipeline-reliability mechanism rather than consumer outcomes, company conduct, regulatory compliance, misconduct, or population prevalence.

\subsection{Evidence conditions}

\subsubsection{Clean}

The clean condition served as the non-degraded experimental baseline. The term clean does not imply that the evidence representation was exhaustive, technically perfect, or guaranteed to produce a parser-valid model output. It identifies the baseline representation against which the three controlled evidence interventions were compared.

\subsubsection{Compressed-lossy}

The compressed-lossy condition represented an evidence state that had been condensed in a way capable of removing details required by downstream stages. The intervention was designed to preserve broad case context while reducing potentially consequential detail.

This condition represents information loss associated with processes such as summarisation, compression, filtering, or transformation before evidence reaches a downstream component.

\subsubsection{Partial-dropout}

The partial-dropout condition represented the controlled absence of required evidence elements. Unlike the noisy-conflicting condition, partial-dropout did not introduce an explicit contradiction.

It modelled a setting in which missing information creates evidential insufficiency or uncertainty without necessarily providing a clear signal that the remaining evidence is unreliable.

\subsubsection{Noisy-conflicting}

The noisy-conflicting condition introduced competing, distracting, or internally difficult-to-reconcile evidence.

This intervention tested the response of the pipeline after evidential conflict had entered the sequence, including whether the conflict was detected and whether subsequent audit and escalation functions produced successful outcomes.

The three degraded conditions do not represent every possible evidence failure. They operationalise three controlled degradation families:

\begin{itemize}
    \item information loss through compression;
    \item information loss through omission; and
    \item evidential conflict through noise.
\end{itemize}

They do not directly model temporal staleness, distribution shift, adversarial manipulation, policy ambiguity, retrieval failure, or delayed ground truth.

\subsection{Pipeline stages}

\subsubsection{Decision}

The decision stage generated an assessment or recommendation from the supplied evidence state. Its principal measurements were:

\begin{itemize}
    \item parser validity;
    \item evidence adequacy as encoded in the sanitised output; and
    \item evidence-sensitive stage success.
\end{itemize}

The decision-stage outputs were evaluated only as computational outputs of the experimental pipeline. They were not treated as real financial decisions or regulatory determinations.

\subsubsection{Audit}

The audit stage examined whether the available evidence and preceding decision state indicated degradation or concern.

Among parser-valid audit rows, two additional indicators were evaluated:

\begin{itemize}
    \item degradation detection; and
    \item audit false assurance.
\end{itemize}

These indicators are computational, model-output-coded variables produced within the experimental pipeline. They do not represent independent institutional auditing, professional assurance, or regulatory review.

\subsubsection{Escalation}

The escalation stage evaluated the downstream response after evidence degradation had entered the sequence.

Among parser-valid escalation rows, the recovery indicator recorded whether the implemented mechanism restored operational stage success. The corresponding failure indicator recorded continued lack of recovery under the committed scoring contract.

Detection and recovery were therefore treated as separate functions. An audit stage could identify an evidential problem while the escalation stage still failed to restore a successful evidence-sensitive outcome.

\subsection{Model execution and accounting}

The experiment used the model identifier GLM-5.2.

The planned and ledgered execution contained 720 calls. The complete ledger contained

\[
470 + 250 = 720
\]

calls, comprising 470 parser-valid and 250 parser-invalid ledger records.

The retained sanitised execution dataset contained 713 rows:

\[
470 + 243 = 713,
\]

comprising 470 parser-valid and 243 parser-invalid retained rows. Every parser-valid ledger output was retained.

The difference between the planned ledger and retained execution data was therefore

\[
720 - 713 = 7.
\]

Seven attempted calls had ledger records but no retained sanitised execution row because of recorded URL or timeout errors. Three of these missing retained observations occurred under the clean condition and four under the noisy-conflicting condition. No compressed-lossy or partial-dropout execution row was missing.

By pipeline stage, the seven missing rows consisted of:

\begin{itemize}
    \item three decision rows;
    \item three audit rows; and
    \item one escalation row.
\end{itemize}

These seven ledger-only failures were excluded from execution-level analyses and were not imputed. Condition-stage rates were calculated using the actual number of retained rows in each cell, producing retained denominators ranging from 58 to 60.

Matched degraded-minus-clean comparisons included only base cases for which both the clean observation and corresponding degraded observation were retained for the same pipeline stage. The resulting paired sample sizes ranged from 57 to 60.

The maximum cumulative estimated model-execution cost recorded during the experiment was USD~2.2731216. The approved execution ceiling was USD~8.00.

\subsection{Uncertainty analysis}

Matched degraded-minus-clean mean differences were evaluated for three model-output-coded measurements:

\begin{itemize}
    \item parser validity;
    \item evidence-sensitive stage success; and
    \item the committed evidence-state-adequacy indicator.
\end{itemize}

Uncertainty was estimated using a nonparametric bootstrap over matched base cases. For each condition-stage-metric comparison, matched case pairs were resampled together so that the relationship between the clean and degraded observations for the same base case was preserved.

Each comparison used:

\begin{itemize}
    \item 2,000 bootstrap resamples; and
    \item fixed random seed 5205.
\end{itemize}

Percentile-based 95\% bootstrap intervals were calculated for all 27 combinations of

\[
3 \text{ degraded conditions}
\times
3 \text{ pipeline stages}
\times
3 \text{ measurements}.
\]

The intervals characterise uncertainty across the selected sanitised base cases under the implemented pairing, execution, and scoring procedures. They are not interpreted as population-level confidence intervals for deployed financial systems, consumers, institutions, or model providers.

The analysis records whether each interval includes zero. It does not use the bootstrap results to make unsupported claims of statistical significance, population generalisability, or causal identification beyond the controlled within-case intervention design.

\subsection{Reproducibility and reporting boundary}

The committed research artifacts support deterministic verification of:

\begin{itemize}
    \item execution and retention counts;
    \item parser accounting;
    \item metric definitions;
    \item condition-stage tables;
    \item matched comparisons;
    \item bootstrap intervals;
    \item cascade classifications;
    \item figure source data; and
    \item claim-to-evidence mappings.
\end{itemize}

This reporting approach is intended to make the analysis inspectable and traceable, consistent with established machine-learning reproducibility principles \citep{pineau2021reproducibility}.

The shared artifacts do not support exact prompt-response replay because raw prompts, raw model responses, raw CFPB records, JSONL execution archives, API credentials, and environment files are excluded from the reporting package.

The study therefore provides artifact-level reproducibility and auditability rather than unrestricted reconstruction of every original hosted-model interaction.

\section{Results}
\label{sec:results}

\subsection{Execution and parser accounting}

The planned execution contained 720 model calls, all of which were represented in the execution ledger. Of these, 713 produced retained sanitised execution rows. Table~\ref{tab:execution-accounting} summarises the accounting.

\begin{table}[H]
\centering
\caption{Execution and parser accounting.}
\label{tab:execution-accounting}
\resizebox{\textwidth}{!}{
\begin{tabular}{lrl}
\toprule
\textbf{Measure} & \textbf{Value} & \textbf{Interpretation} \\
\midrule
Planned calls & 720 & \(60 \times 4 \times 3\) design \\
Ledger rows & 720 & Complete plan-level accounting \\
Retained execution rows & 713 & Rows used for execution-level metrics \\
Ledger parser-valid rows & 470 & Structurally valid ledger outputs \\
Ledger parser-invalid rows & 250 & Includes seven ledger-only failures \\
Retained parser-valid rows & 470 & Every parser-valid ledger row retained \\
Retained parser-invalid rows & 243 & Parser-invalid retained outputs \\
Ledger-only missing rows & 7 & Excluded rather than imputed \\
Maximum estimated cost & USD 2.2731216 & Below approved USD 8.00 ceiling \\
\bottomrule
\end{tabular}
}
\end{table}

The seven ledger-only observations were not evenly distributed. Three occurred under the clean condition and four under noisy-conflicting evidence. By stage, the missing observations consisted of three decision rows, three audit rows, and one escalation row. No compressed-lossy or partial-dropout execution row was missing. Retained condition-stage denominators therefore ranged from 58 to 60.

\subsection{Parser validity and stage success by condition}

Table~\ref{tab:condition-stage-results} reports the retained observations, parser-valid outputs, and evidence-sensitive stage-success outcomes for every condition-stage combination.

\begin{table}[H]
\centering
\caption{Condition-by-stage structural and evidence-sensitive outcomes.}
\label{tab:condition-stage-results}
\resizebox{\textwidth}{!}{
\begin{tabular}{llrrrrr}
\toprule
\textbf{Condition} &
\textbf{Stage} &
\textbf{Retained \(n\)} &
\textbf{Parser-valid \(n\)} &
\textbf{Parser-valid rate} &
\textbf{Stage-success \(n\)} &
\textbf{Stage-success rate} \\
\midrule
Clean & Decision & 59 & 30 & 0.508475 & 30 & 0.508475 \\
Clean & Audit & 58 & 30 & 0.517241 & 30 & 0.517241 \\
Clean & Escalation & 60 & 25 & 0.416667 & 25 & 0.416667 \\
\midrule
Compressed-lossy & Decision & 60 & 51 & 0.850000 & 0 & 0.000000 \\
Compressed-lossy & Audit & 60 & 48 & 0.800000 & 0 & 0.000000 \\
Compressed-lossy & Escalation & 60 & 43 & 0.716667 & 0 & 0.000000 \\
\midrule
Partial-dropout & Decision & 60 & 35 & 0.583333 & 0 & 0.000000 \\
Partial-dropout & Audit & 60 & 41 & 0.683333 & 0 & 0.000000 \\
Partial-dropout & Escalation & 60 & 32 & 0.533333 & 0 & 0.000000 \\
\midrule
Noisy-conflicting & Decision & 58 & 46 & 0.793103 & 0 & 0.000000 \\
Noisy-conflicting & Audit & 59 & 52 & 0.881356 & 0 & 0.000000 \\
Noisy-conflicting & Escalation & 59 & 37 & 0.627119 & 0 & 0.000000 \\
\bottomrule
\end{tabular}
}
\end{table}

The clean-condition parser-validity rates were 0.508475 at decision, 0.517241 at audit, and 0.416667 at escalation. The clean condition therefore served as an evidence baseline rather than a technically perfect output condition.

Every degraded condition-stage cell had a higher parser-validity rate than its corresponding clean cell. At the same time, every degraded condition-stage cell had an operational stage-success rate of 0.0.

These cell-wise comparisons are descriptive. The matched analysis below preserves base-case identity and therefore provides the primary basis for evaluating degraded-minus-clean changes. Because only cases retained under both conditions enter each matched comparison, the paired means differ slightly from the unpaired cell-rate differences in Table~\ref{tab:condition-stage-results}.

Across the nine paired degraded comparisons, parser-validity mean changes ranged from \(+0.067797\) to \(+0.368421\), whereas operational stage-success mean changes ranged from \(-0.517241\) to \(-0.406780\). The two measurements therefore moved in opposite directions under every evaluated degraded condition-stage comparison.

\begin{figure}[H]
    \centering
    \includegraphics[width=0.95\textwidth]{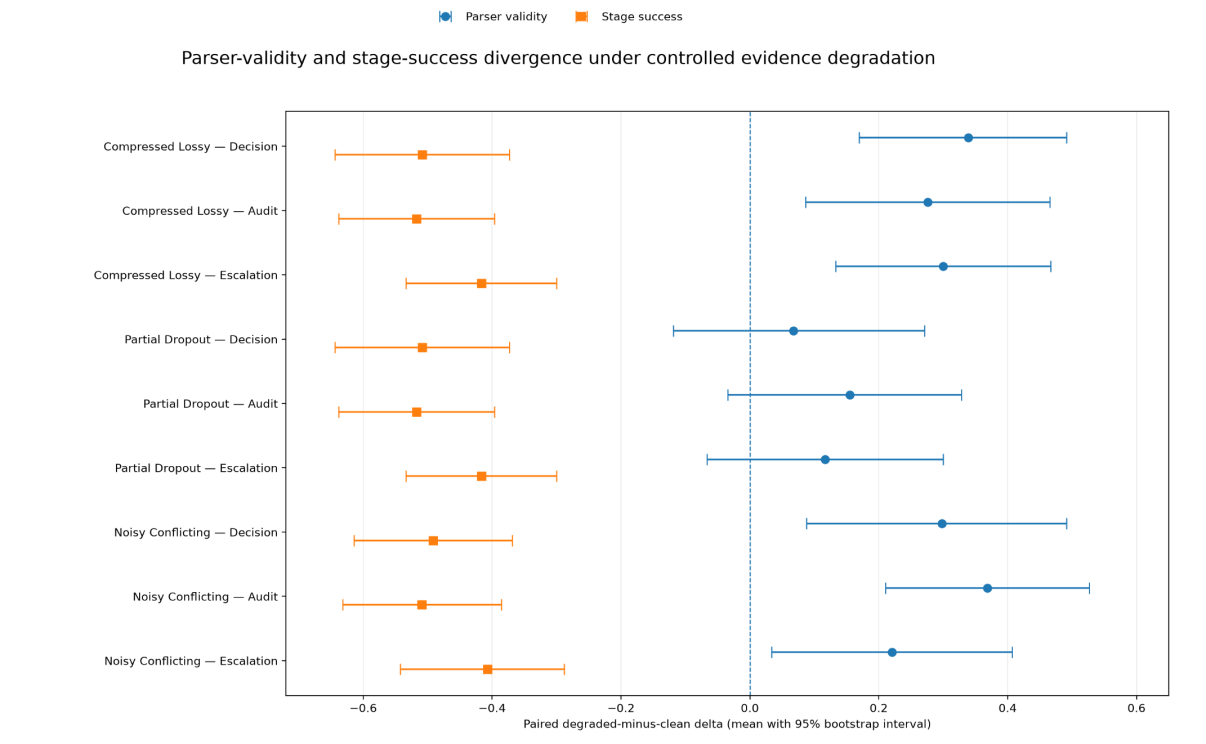}
    \caption{Paired parser-validity and evidence-sensitive stage-success changes under controlled degradation. Points show degraded-minus-clean means and 95\% bootstrap intervals from 2,000 resamples with fixed seed 5205. All nine parser-validity point estimates are positive, although the three partial-dropout intervals include zero. All nine stage-success intervals remain below zero.}
    \label{fig:parser-stage-divergence}
\end{figure}

\subsection{Paired bootstrap results}

Tables~\ref{tab:bootstrap-pv}--\ref{tab:bootstrap-adequacy} report the matched degraded-minus-clean estimates and percentile-based 95\% bootstrap intervals.

\begin{table}[H]
\centering
\caption{Paired degraded-minus-clean changes in parser validity.}
\label{tab:bootstrap-pv}
\resizebox{\textwidth}{!}{
\begin{tabular}{llrrrr}
\toprule
\textbf{Stage} &
\textbf{Degraded condition} &
\textbf{Paired \(n\)} &
\textbf{Mean \(\Delta\)} &
\textbf{95\% low} &
\textbf{95\% high} \\
\midrule
Decision & Compressed-lossy & 59 & +0.338983 & +0.169492 & +0.491525 \\
Decision & Partial-dropout & 59 & +0.067797 & -0.118644 & +0.271186 \\
Decision & Noisy-conflicting & 57 & +0.298246 & +0.087719 & +0.491228 \\
Audit & Compressed-lossy & 58 & +0.275862 & +0.086207 & +0.465517 \\
Audit & Partial-dropout & 58 & +0.155172 & -0.034483 & +0.328017 \\
Audit & Noisy-conflicting & 57 & +0.368421 & +0.210526 & +0.526316 \\
Escalation & Compressed-lossy & 60 & +0.300000 & +0.133333 & +0.466667 \\
Escalation & Partial-dropout & 60 & +0.116667 & -0.066667 & +0.300000 \\
Escalation & Noisy-conflicting & 59 & +0.220339 & +0.033898 & +0.406780 \\
\bottomrule
\end{tabular}
}
\end{table}

All nine parser-validity point estimates were positive. Six of the nine bootstrap intervals remained entirely above zero. The three intervals containing zero were all partial-dropout comparisons:

\begin{itemize}
    \item decision: \([-0.118644,\ 0.271186]\);
    \item audit: \([-0.034483,\ 0.328017]\); and
    \item escalation: \([-0.066667,\ 0.300000]\).
\end{itemize}

The parser-validity results therefore support directional improvement in every paired comparison, but not uniform interval separation from zero under partial dropout.

\begin{table}[H]
\centering
\caption{Paired degraded-minus-clean changes in operational stage success.}
\label{tab:bootstrap-success}
\resizebox{\textwidth}{!}{
\begin{tabular}{llrrrr}
\toprule
\textbf{Stage} &
\textbf{Degraded condition} &
\textbf{Paired \(n\)} &
\textbf{Mean \(\Delta\)} &
\textbf{95\% low} &
\textbf{95\% high} \\
\midrule
Decision & Compressed-lossy & 59 & -0.508475 & -0.644068 & -0.372881 \\
Decision & Partial-dropout & 59 & -0.508475 & -0.644068 & -0.372881 \\
Decision & Noisy-conflicting & 57 & -0.491228 & -0.614035 & -0.368421 \\
Audit & Compressed-lossy & 58 & -0.517241 & -0.637931 & -0.396121 \\
Audit & Partial-dropout & 58 & -0.517241 & -0.637931 & -0.396121 \\
Audit & Noisy-conflicting & 57 & -0.508772 & -0.631579 & -0.385965 \\
Escalation & Compressed-lossy & 60 & -0.416667 & -0.533333 & -0.300000 \\
Escalation & Partial-dropout & 60 & -0.416667 & -0.533333 & -0.300000 \\
Escalation & Noisy-conflicting & 59 & -0.406780 & -0.542373 & -0.288136 \\
\bottomrule
\end{tabular}
}
\end{table}

All nine operational stage-success estimates were negative, and all nine corresponding bootstrap intervals remained below zero.

\begin{table}[H]
\centering
\caption{Paired degraded-minus-clean changes in the model-output-coded evidence-state-adequacy indicator.}
\label{tab:bootstrap-adequacy}
\resizebox{\textwidth}{!}{
\begin{tabular}{llrrrr}
\toprule
\textbf{Stage} &
\textbf{Degraded condition} &
\textbf{Paired \(n\)} &
\textbf{Mean \(\Delta\)} &
\textbf{95\% low} &
\textbf{95\% high} \\
\midrule
Decision & Compressed-lossy & 59 & -0.508475 & -0.644068 & -0.372881 \\
Decision & Partial-dropout & 59 & -0.508475 & -0.644068 & -0.372881 \\
Decision & Noisy-conflicting & 57 & -0.473684 & -0.596491 & -0.350877 \\
Audit & Compressed-lossy & 58 & -0.517241 & -0.637931 & -0.396121 \\
Audit & Partial-dropout & 58 & -0.517241 & -0.637931 & -0.396121 \\
Audit & Noisy-conflicting & 57 & -0.508772 & -0.631579 & -0.385965 \\
Escalation & Compressed-lossy & 60 & -0.416667 & -0.533333 & -0.300000 \\
Escalation & Partial-dropout & 60 & -0.416667 & -0.533333 & -0.300000 \\
Escalation & Noisy-conflicting & 59 & -0.406780 & -0.542373 & -0.288136 \\
\bottomrule
\end{tabular}
}
\end{table}

All nine evidence-state-adequacy estimates were negative, and all nine corresponding intervals remained below zero. This provides a second model-output-coded indicator that changed in the same direction as operational stage success under the tested degradations.

Together, the paired results show the central reliability-layer divergence observed in this experiment: parser-validity point estimates moved in a favourable direction while the two evidence-sensitive indicators moved in the opposite direction.

\subsection{Audit detection and false assurance}

Among parser-valid degraded audit outputs, the degradation-detection indicator was positive in every evaluated row:

\begin{itemize}
    \item compressed-lossy: 48 of 48, or 1.0;
    \item partial-dropout: 41 of 41, or 1.0; and
    \item noisy-conflicting: 52 of 52, or 1.0.
\end{itemize}

The implemented audit field therefore consistently signalled degradation among parser-valid degraded audit outputs.

Detection did not eliminate the audit false-assurance indicator. It appeared in:

\begin{itemize}
    \item 3 of 48 compressed-lossy parser-valid audit rows, or 0.062500;
    \item 2 of 41 partial-dropout parser-valid audit rows, or 0.048780; and
    \item 1 of 52 noisy-conflicting parser-valid audit rows, or 0.019231.
\end{itemize}

These rates were non-zero despite degradation detection being 1.0 in each degraded condition. Detection, false assurance, and downstream recovery are therefore reported as distinct outcomes. Both audit detection and audit false assurance remain model-output-coded computational indicators rather than independent external audit findings.

\begin{figure}[H]
    \centering
    \includegraphics[width=0.82\textwidth]{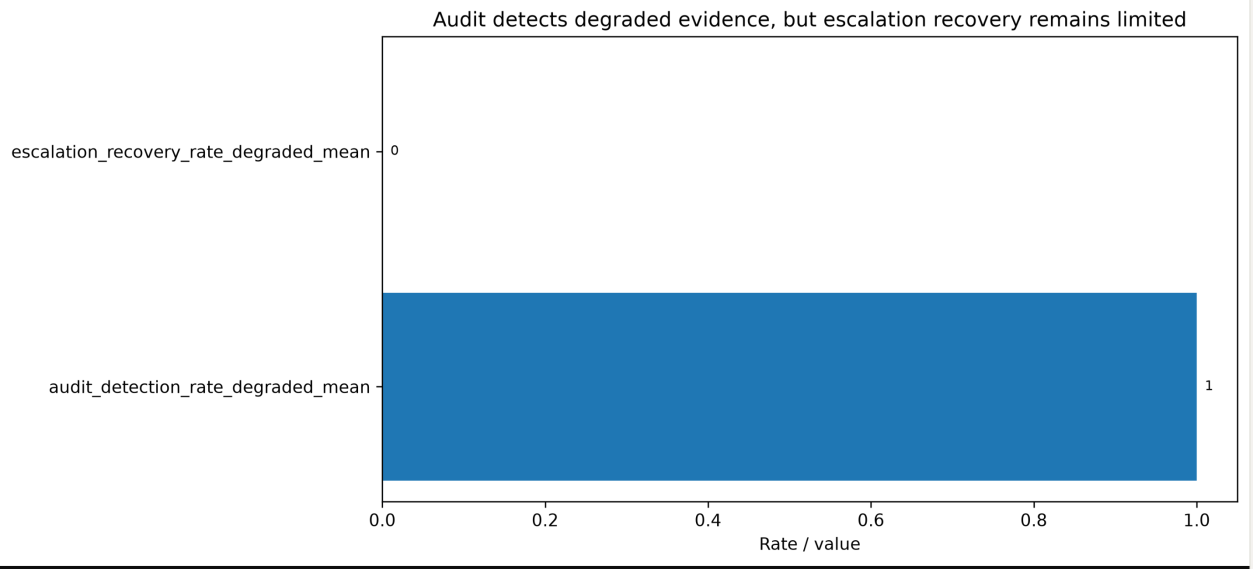}
    \caption{Audit detection and escalation recovery under controlled evidence degradation. Among parser-valid degraded outputs, the audit degradation-detection rate was 1.0 in each degraded condition, while escalation recovery was 0.0 in every degraded condition. These are model-output-coded indicators from the implemented computational pipeline and do not represent institutional audit effectiveness.}
    \label{fig:audit-recovery}
\end{figure}

\subsection{Escalation recovery}

In the clean condition, all 25 parser-valid escalation outputs were stage-successful, giving a success rate of 1.0 when the denominator was restricted to parser-valid clean escalation outputs.

Under degradation, no parser-valid escalation output was coded as recovered:

\begin{itemize}
    \item compressed-lossy: 0 of 43;
    \item partial-dropout: 0 of 32; and
    \item noisy-conflicting: 0 of 37.
\end{itemize}

Escalation recovery was therefore 0.0 in each degraded condition under the implemented scoring contract.

This result establishes detection without recovery within the evaluated pipeline. It does not establish that escalation mechanisms are inherently ineffective or that recovery would remain absent under a different model, prompt, evidence source, retrieval mechanism, human-review pathway, or recovery architecture.

\subsection{Sequence-level reliability cascades}

The sequence-level analysis formed 240 groups, corresponding to 60 base cases under four evidence conditions. Of these, 234 contained all three retained stage rows and six were incomplete.

A total of 223 groups met the operational cascade-failure definition, while 17 were classified as preserved successes. The aggregate cascade-failure rate was therefore

\[
\frac{223}{240} = 0.929167.
\]

This statistic includes both clean and degraded conditions and is not interpreted as a degraded-condition treatment effect or as an estimate of failure prevalence in deployment.

\begin{figure}[H]
    \centering
    \includegraphics[width=0.88\textwidth]{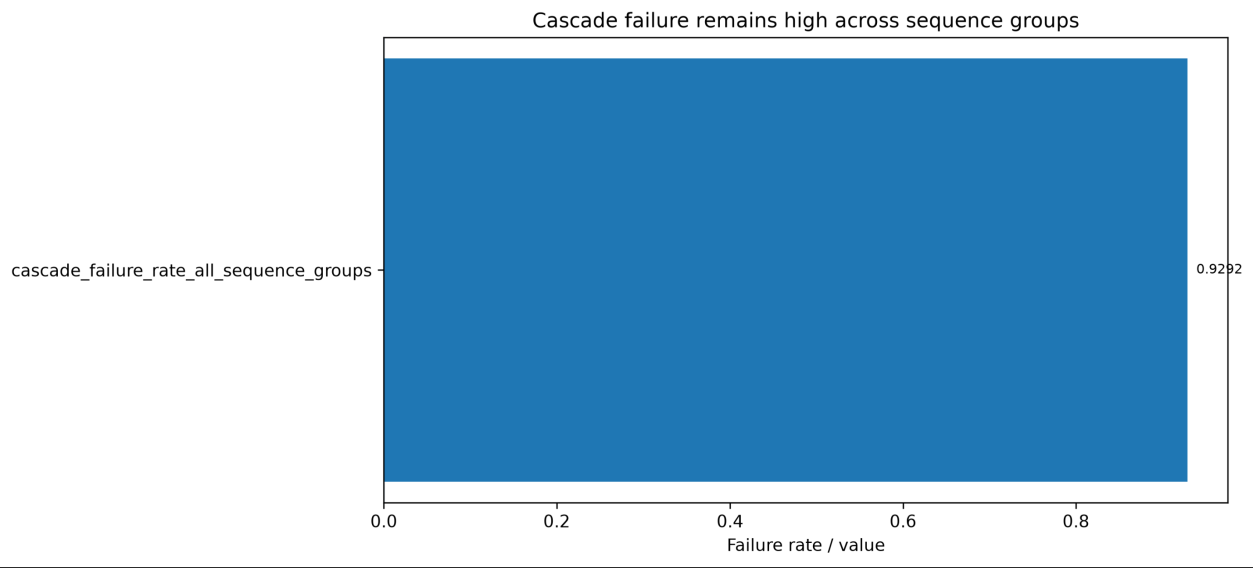}
  \caption{Sequence-level cascade-failure rate across all evidence conditions. The operational cascade definition classified 223 of 240 condition-linked sequence groups as failures. This aggregate includes both clean and degraded conditions and should not be interpreted as a degradation treatment effect or as an estimate of failure prevalence in deployed systems.}
\label{fig:cascade-rate}
\end{figure}

Table~\ref{tab:cascade-composition} reports the sequence-level composition.

\begin{table}[H]
\centering
\caption{Sequence-level reliability-cascade composition.}
\label{tab:cascade-composition}
\resizebox{\textwidth}{!}{
\begin{tabular}{lrrrrl}
\toprule
\textbf{Condition} &
\textbf{Groups} &
\textbf{Complete} &
\textbf{Failures} &
\textbf{Failure rate} &
\textbf{Pattern composition} \\
\midrule
Clean &
60 & 57 & 43 & 0.716667 &
40 parser failure; 3 incomplete; 17 preserved success \\
Compressed-lossy &
60 & 60 & 60 & 1.000000 &
29 parser failure; 29 detected but not recovered; 2 audit false assurance \\
Partial-dropout &
60 & 60 & 60 & 1.000000 &
46 parser failure; 14 detected but not recovered \\
Noisy-conflicting &
60 & 57 & 60 & 1.000000 &
28 parser failure; 28 detected but not recovered; 1 audit false assurance; 3 incomplete \\
\midrule
All conditions &
240 & 234 & 223 & 0.929167 &
143 parser failure; 71 detected but not recovered; 3 audit false assurance; 6 incomplete; 17 preserved success \\
\bottomrule
\end{tabular}
}
\end{table}

Across all 240 sequence groups, the mutually exclusive taxonomy contained:

\begin{itemize}
    \item 143 parser-failure cascades;
    \item 71 detected-but-unrecovered patterns;
    \item 3 audit-false-assurance patterns;
    \item 6 incomplete sequences; and
    \item 17 preserved successes.
\end{itemize}

The five categories sum to 240.

\begin{figure}[H]
    \centering
    \includegraphics[width=0.90\textwidth]{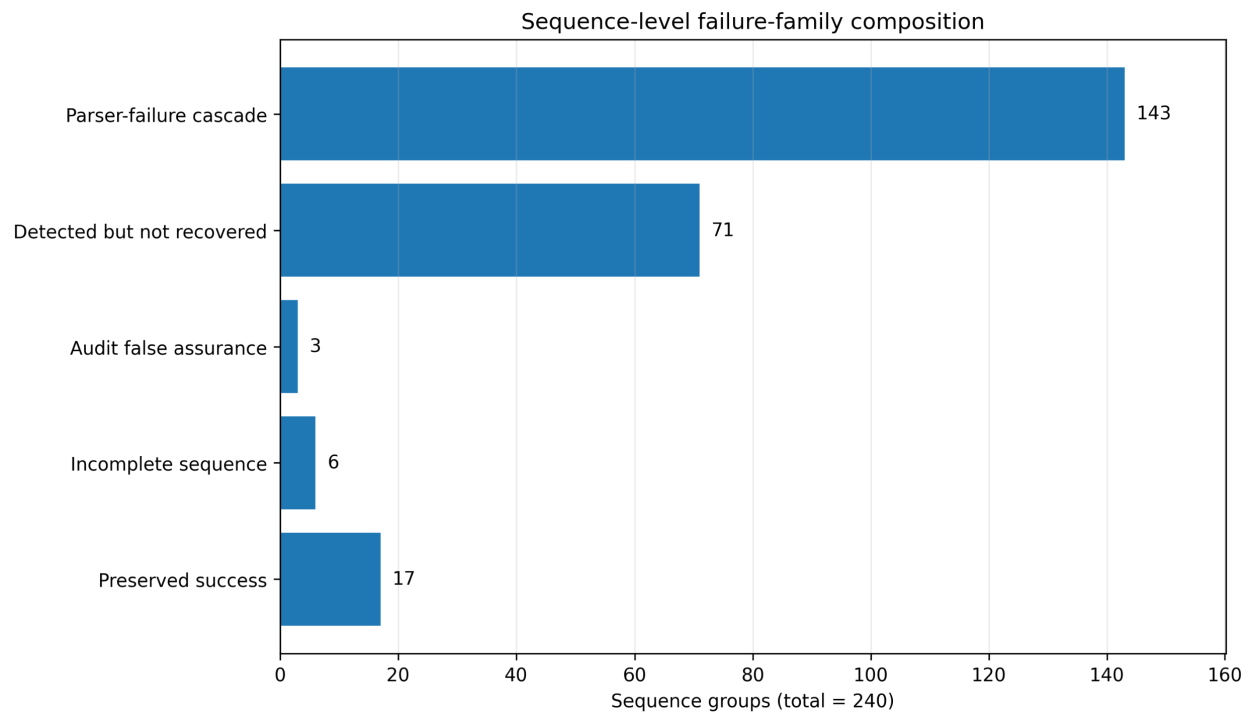}
    \caption{Sequence-level failure-family composition. Across 240 condition-linked groups, the operational taxonomy contained 143 parser-failure cascades, 71 detected-but-unrecovered patterns, three audit-false-assurance patterns, six incomplete sequences, and 17 preserved successes.}
    \label{fig:cascade-composition}
\end{figure}

Clean evidence accounted for 43 cascade failures among 60 groups, including 40 parser-failure cascades and three incomplete sequences; the remaining 17 clean groups were preserved successes. Each degraded condition produced 60 groups meeting the operational cascade-failure definition.

The aggregate cascade statistic serves a diagnostic purpose that differs from the paired degradation analysis. The matched condition-stage comparisons quantify changes associated with controlled evidence degradation, whereas the cascade taxonomy describes the sequence-level outcome family observed for each condition-linked group.

\section{Discussion}
\label{sec:discussion}

\subsection{Reliability-layer divergence}

The central result is not simply that controlled evidence degradation reduced operational stage success. The more informative finding is that structural and evidence-sensitive indicators moved in opposite directions under the same interventions.

Across all nine degraded condition-stage comparisons, parser-validity point estimates were positive, while operational stage-success estimates were negative. All nine stage-success bootstrap intervals remained below zero. By contrast, three parser-validity intervals, all under partial dropout, included zero. The observed pattern therefore supports directional reliability-layer divergence across all tested comparisons, while providing stronger interval separation for some degradation conditions than others.

This distinction matters because a system evaluated only through schema compliance could appear structurally improved even while its evidence-sensitive performance deteriorated. In the evaluated pipeline, parser validity and operational ESR therefore measured different properties of system behaviour rather than interchangeable notions of reliability.

The experiment does not establish why this divergence occurred. Compression or omission might reduce the amount or complexity of information that must be expressed within a structured response, while conflicting evidence might alter response behaviour in ways that affect parsing. These possibilities are interpretations rather than demonstrated mechanisms. The present design establishes the divergence itself, not its causal explanation.

Parser validity nevertheless remains necessary. Structural failure was the largest sequence-level failure family, and automated pipelines require outputs that downstream components can parse reliably. The result therefore does not support replacing structural validation. It supports evaluating structural conformance alongside evidence-sensitive measures rather than treating parser validity as a sufficient proxy for substantive reliability.

\subsection{Interpreting the clean baseline}

The clean evidence condition was not a technically easy output condition. Parser-validity rates were 0.508475 at decision, 0.517241 at audit, and 0.416667 at escalation.

These relatively low baseline rates created substantial headroom for positive parser-validity changes under degraded evidence. The observed positive parser-validity estimates should therefore not be interpreted as evidence that degradation intrinsically improves structural conformance.

The paired design still provides a meaningful within-case comparison because each degraded observation is matched to the clean observation for the same base case and stage. However, the clean baseline constrains the scope of interpretation. The supported result is that parser validity increased directionally relative to this particular baseline while evidence-sensitive stage success deteriorated.

Future replication should test whether the same divergence persists when the clean condition already has high structural validity, for example under different output contracts, schema-constrained decoding, prompting strategies, or model configurations.

\subsection{Interpreting zero degraded stage success}

Every degraded condition-stage cell had an operational stage-success rate of 0.0 under the committed scoring contract. Consistently, all nine paired stage-success bootstrap intervals remained below zero.

This is a strong within-experiment pattern, but the operational definition requires careful interpretation. Stage success required both a parser-valid output and a positive model-output-coded \texttt{validity\_judgment}. Rows without a parser-valid output or a positive validity judgment were not counted as successful.

Operational stage success is therefore an indicator of ESR within this experiment rather than ESR itself. ESR is the broader construct concerning whether an evidence state remains sufficiently complete, grounded, internally consistent, and usable for the function assigned to a stage.

The separately recorded \texttt{evidence\_state\_adequate} variable provides convergent support because all nine corresponding degraded-minus-clean intervals also remained below zero. However, both stage success and evidence-state adequacy are derived from model-output-coded fields rather than independent human or expert adjudication.

The supported conclusion is therefore that multiple committed indicators changed consistently under controlled degradation within the evaluated pipeline. The experiment does not establish that these variables provide an exhaustive or universally valid measurement of Evidence-State Reliability.

Independent adjudication would be required to determine how closely the model-output-coded variables correspond to external assessments of evidence sufficiency, grounding, decision support, audit correctness, and successful recovery.

\subsection{Detection is not recovery}

The audit and escalation results reveal a second important separation between reliability functions.

Among parser-valid degraded audit outputs, the degradation-detection indicator was positive in every evaluated row. Despite this detection rate of 1.0, audit false-assurance indicators remained non-zero, and no parser-valid degraded escalation output was coded as recovered.

Recognition of a degraded evidence state therefore did not imply restoration of successful operation in the implemented pipeline. A monitoring component may identify a problem without having access to the evidence, authority, or mechanism required to resolve it.

This distinction is important for interpreting Figure~\ref{fig:audit-recovery}. A high detection rate alone does not characterise the success of an audit-and-escalation pathway. Evaluation should distinguish at least:

\begin{itemize}
    \item degradation detection;
    \item false assurance;
    \item escalation action;
    \item recovery;
    \item safe deferral; and
    \item unresolved failure.
\end{itemize}

Effective recovery may require information or capabilities that are not available to a monitoring stage, such as retrieving missing evidence, resolving contradictory information, applying a different decision policy, involving an authorised reviewer, or declining to proceed.

The present experiment establishes that recovery was absent under the implemented escalation mechanism. It does not compare alternative recovery architectures or establish that recovery is impossible under other designs.

\subsection{Why sequence-aware evaluation matters}

The sequence-level analysis shows why stage-level metrics and endpoint metrics answer different questions.

Across the 240 condition-linked groups, the operational taxonomy contained 143 parser-failure cascades, 71 detected-but-unrecovered patterns, three audit-false-assurance patterns, six incomplete sequences, and 17 preserved successes.

These categories represent different technical problems. Parser-failure cascades concern structural conformance or validation. Detected-but-unrecovered degradation indicates that recognition was not accompanied by successful remediation. Audit false assurance identifies sequences containing the committed false-assurance indicator. Incomplete sequences reflect execution or persistence problems rather than only model reasoning.

A single end-to-end failure statistic would obscure these distinctions.

The aggregate cascade-failure rate of 0.929167 should also be interpreted carefully. It includes a clean-condition failure rate of 0.716667 and is therefore not a pure degradation effect. The paired condition-stage comparisons provide the stronger basis for interpreting changes associated with controlled evidence degradation, whereas the cascade taxonomy provides a diagnostic description of the observed sequence outcomes.

Taken together, the results support a layered approach to evaluating multi-stage LLM systems. Structural validity, evidence-state adequacy, evidence-sensitive stage success, degradation detection, false assurance, recovery, and sequence completion should be measured separately when they represent different system functions.

The same principle has implications for pipeline design. A structural contract specifies the format required for automated consumption, but an evidence-sensitive system also requires rules governing what evidence is sufficient, how conflict or absence should be represented, when deferral should occur, and what changes after a problem has been detected.

Without such distinctions, a pipeline can remain structurally operational while the evidence available to support its substantive function has deteriorated. The present results therefore support treating intermediate evidence states and stage transitions as measurable parts of reliability evaluation rather than evaluating only the final response.

\section{Threats to Validity and Limitations}
\label{sec:limitations}

The limitations of the study are considered across construct, internal, external, conclusion, measurement-coverage, and reproducibility dimensions, following established empirical-research validity principles \citep{wohlin2012experimentation}. These limitations constrain the interpretation and generalisability of the findings but do not alter the reported within-design comparisons.

\subsection{Construct validity}

Evidence-State Reliability is broader than the operational variables used in this experiment.

The primary operational ESR indicator, stage success, required both:

\begin{enumerate}
    \item a parser-valid output; and
    \item a positive model-output-coded \texttt{validity\_judgment}.
\end{enumerate}

Rows without a parser-valid output or a positive validity judgment were not counted as successful. This rule was applied consistently, but it may combine genuine evidence inadequacy with behaviour induced by the model, prompt, or output contract.

The additional indicators for evidence-state adequacy, audit detection, audit false assurance, and escalation recovery were likewise derived from sanitised model-output fields rather than independent human, expert, or rule-based adjudication.

Agreement among these indicators provides convergent evidence within the implemented experiment. It does not establish that they exhaustively represent ESR or that the same operational definitions would remain valid across different models, prompts, domains, or scoring contracts.

Future validation should compare these model-output-coded variables with independent assessments of evidence sufficiency, grounding, consistency, audit correctness, appropriate deferral, and successful recovery.

\subsection{Internal validity}

The paired design controls for base-case identity by comparing clean and degraded observations for the same case and pipeline stage. Nevertheless, the observed reliability-layer divergence may also depend on properties of the experimental implementation, including:

\begin{itemize}
    \item prompt design;
    \item parser-contract difficulty;
    \item output schema;
    \item pipeline ordering;
    \item stage-specific instructions;
    \item scoring rules; and
    \item interactions among these elements.
\end{itemize}

The clean condition had relatively low parser-validity rates, ranging from approximately 0.42 to 0.52 across stages. This created substantial headroom for positive parser-validity changes under degraded conditions. The results therefore do not establish that evidence degradation intrinsically improves structural validity.

The primary experiment also consisted of one scaled execution using the evaluated model configuration. It therefore does not quantify run-to-run variability under repeated executions of the same configuration.

Seven of the 720 ledgered calls had no retained sanitised execution row because of recorded URL or timeout errors. These observations were excluded rather than imputed. Every parser-valid ledger output was retained, but the execution dataset was not perfectly complete.

Matched analyses reduced the effect of these missing observations by including only base cases for which both clean and degraded observations were retained for the relevant stage. Even so, paired sample sizes varied from 57 to 60, and the missing observations could have affected individual estimates.

\subsection{External validity}

The primary study evaluated:

\begin{itemize}
    \item one GLM-5.2 model configuration;
    \item one decision--audit--escalation pipeline;
    \item one scaled execution run;
    \item one selected set of 60 sanitised base cases;
    \item one committed scoring contract; and
    \item three controlled evidence-degradation families.
\end{itemize}

The findings therefore do not establish cross-model generality, provider independence, domain independence, or universal behaviour across multi-stage LLM systems.

The 60 selected cases are not claimed to constitute a representative or probability sample of consumer complaints, financial decisions, or deployed AI interactions.

CFPB provenance provides a realistic complaint-context substrate, but it does not establish complaint truth, institutional misconduct, regulatory violations, consumer-harm prevalence, population-level frequencies, or the quality of actual financial decisions.

The tested interventions represent information loss through compression, information loss through omission, and evidential conflict through noise. They do not directly evaluate other reliability threats such as:

\begin{itemize}
    \item temporal staleness;
    \item distribution shift;
    \item retrieval failure;
    \item adversarial manipulation;
    \item policy ambiguity;
    \item delayed ground truth;
    \item malicious evidence injection; or
    \item changes in model or provider behaviour.
\end{itemize}

Replication across models, domains, output contracts, evidence sources, and recovery architectures is therefore required before broader generalisation.

\subsection{Conclusion validity}

The paired bootstrap intervals characterise uncertainty across the selected sanitised base cases under the implemented model configuration, matching procedure, and scoring contract.

All nine operational stage-success intervals remained below zero. In contrast, three parser-validity intervals---the partial-dropout comparisons at decision, audit, and escalation---included zero.

The evidence therefore supports consistent negative changes in operational stage success within the evaluated design. It does not support a claim of uniformly interval-separated parser-validity improvement under every degraded condition.

The bootstrap intervals should not be interpreted as population-level confidence intervals for deployed financial systems, consumers, organisations, or model providers. They describe variability across the selected matched cases rather than sampling uncertainty from a defined real-world population.

The aggregate cascade-failure rate also requires separate interpretation. Because it includes clean-condition failures, it is not:

\begin{itemize}
    \item a pure degradation effect;
    \item a deployment failure probability;
    \item an estimate of consumer risk; or
    \item a prevalence estimate for real-world AI systems.
\end{itemize}

The paired condition-stage comparisons provide the stronger basis for interpreting changes associated with controlled evidence degradation. The cascade statistic serves a different purpose by describing the sequence-level outcome families observed in the experimental run.

\subsection{Measurement coverage}

The experiment did not directly measure:

\begin{itemize}
    \item latency;
    \item condition-specific cost variation;
    \item calibration;
    \item confidence quality;
    \item human interpretability;
    \item explanation faithfulness;
    \item downstream user reliance;
    \item user harm;
    \item institutional decision quality;
    \item alternative retrieval policies;
    \item safe-deferral quality;
    \item human escalation performance; or
    \item competing recovery architectures.
\end{itemize}

These properties may interact with Evidence-State Reliability but remain distinct evaluation targets.

The three controlled degradation conditions also should not be interpreted as complete reproductions of naturally occurring compression, omission, or conflict processes. The experiment establishes behaviour under defined interventions; it does not establish how frequently or severely equivalent evidence failures occur in deployed systems.

Similarly, the zero-recovery result applies only to the implemented escalation mechanism. The experiment did not test whether additional retrieval, another model, a revised policy, or a human reviewer would have restored successful operation.

\subsection{Reproducibility limitations}

The committed sanitised artifacts support deterministic verification of execution and retention counts, parser accounting, metric calculations, condition-stage tables, matched comparisons, bootstrap intervals, cascade classifications, figure source values, and claim-to-evidence mappings.

However, raw prompts, raw model responses, raw CFPB records, JSONL execution archives, API credentials, and environment files were excluded from the reporting package by design.

Exact prompt-response replay and complete reconstruction of every original model interaction are therefore not claimed. The available artifacts support analysis-level reproducibility and auditability rather than unrestricted computational replication.

A future replication using the same model identifier may also produce different outputs if the hosted model, provider infrastructure, or inference environment changes over time.

Finally, the audit stage evaluated here is computational rather than institutional. The observed zero recovery rate applies to the implemented escalation mechanism and does not establish that recovery is impossible under alternative retrieval, deferral, human-review, or decision-authority arrangements.

\section{Ethical and Data-Use Considerations}
\label{sec:ethics}

The study used sanitised derivative evidence packets constructed from the public context of the Consumer Financial Protection Bureau Consumer Complaint Database. These materials were used solely to evaluate a controlled reliability mechanism within a multi-stage language-model pipeline.

The experiment did not evaluate live consumer decisions and did not produce actual credit, lending, regulatory, legal, or financial determinations. No experimental output was used to affect an individual, organisation, account, complaint, or institutional process.

CFPB provenance was used to provide a realistic complaint-context substrate. It was not used to establish:

\begin{itemize}
    \item whether an individual complaint was factually correct;
    \item whether a named company engaged in misconduct;
    \item whether a legal or regulatory violation occurred;
    \item whether a consumer experienced harm;
    \item whether a financial decision was appropriate; or
    \item how frequently an observed failure pattern would occur in deployment.
\end{itemize}

Consumer complaints are therefore not treated as adjudicated findings or as a representative sample of consumer experience. The study does not draw conclusions about individual consumers, specific organisations, market prevalence, regulatory compliance, or the safety of deployed financial systems.

The shared research boundary excluded:

\begin{itemize}
    \item raw CFPB records;
    \item direct personal identifiers;
    \item raw prompts;
    \item raw model responses;
    \item JSONL execution archives;
    \item API credentials; and
    \item environment files.
\end{itemize}

Only sanitised, derived, and aggregate research artifacts were retained within the reported evidence package.

The decision, audit, and escalation outputs analysed in this study are computational and model-output-coded. They do not constitute professional financial advice, institutional auditing, regulatory assurance, or human expert adjudication.

The findings should therefore be interpreted strictly as evidence about the behaviour of the evaluated experimental pipeline under controlled evidence-state interventions. They should not be used to make decisions or claims concerning individual consumers, named organisations, actual complaints, or the safety and compliance of deployed financial systems.

\section{Reproducibility and Availability}
\label{sec:reproducibility}

The code, sanitised evidence-state contracts, execution accounting, sanitised execution summaries, condition-stage tables, matched-comparison outputs, bootstrap intervals, audit and escalation metrics, cascade-sequence classifications, figure source data, validation reports, and claim-traceability materials are available in the project repository:

\begin{center}
\url{https://github.com/NaimurRahmanR/evidence-state-reliability}
\end{center}

The principal empirical analysis was generated from the following repository checkpoint:

\begin{center}
\texttt{c3e802c71976faae34ac3f327b537e12916bc970}
\end{center}

The final manuscript and submission-gate materials were prepared at the following repository checkpoint:

\begin{center}
\texttt{3144d46e3c59b011f2939d7ab3a590c688543492}
\end{center}

Reporting both checkpoints distinguishes the evidence state used for the principal analysis from the later repository state used for manuscript preparation and submission validation.

The shared artifacts support deterministic verification of:

\begin{itemize}
    \item planned, ledgered, and retained execution counts;
    \item parser-valid and parser-invalid accounting;
    \item metric definitions and formulas;
    \item condition-stage estimates;
    \item matched degraded-minus-clean comparisons;
    \item bootstrap confidence intervals;
    \item audit-detection and false-assurance results;
    \item escalation-recovery results;
    \item cascade classifications;
    \item figure source values; and
    \item mappings between reported claims and supporting artifacts.
\end{itemize}

The repository does not provide unrestricted reconstruction of every original model interaction. Raw CFPB records, raw prompts, raw model responses, JSONL execution archives, API credentials, and environment files are excluded from the shared reporting boundary.

Accordingly, the research package supports artifact-level reproducibility, traceability, and independent verification of the reported analysis. It does not support exact prompt-response replay or complete computational replication of the original hosted-model execution.

\section{Conclusion}
\label{sec:conclusion}

This study introduced and operationalised Evidence-State Reliability as a distinct evaluation layer for multi-stage LLM pipelines. The framework evaluates structural parser validity alongside evidence-sensitive stage success, controlled evidence-state interventions, matched uncertainty analysis, audit detection, false assurance, escalation recovery, and sequence-level reliability patterns.

The framework was evaluated in one scaled GLM-5.2 experiment using 60 sanitised base cases represented under clean, compressed-lossy, partial-dropout, and noisy-conflicting evidence conditions. Each condition was processed through decision, audit, and escalation stages, producing 720 planned and ledgered model calls and 713 retained sanitised execution rows.

The primary empirical finding was a consistent divergence between structural and evidence-sensitive measurements. Across all nine matched degraded-minus-clean condition-stage comparisons, operational stage-success estimates were negative and all corresponding 95\% bootstrap intervals remained below zero. All nine parser-validity point estimates were positive, although the three partial-dropout parser-validity intervals included zero. Within the evaluated pipeline, structural conformance could therefore move in a favourable direction while evidence-sensitive stage success deteriorated under the same controlled intervention.

The audit and escalation results revealed a second distinction. Among parser-valid degraded audit outputs, the degradation-detection indicator was 1.0 in each degraded condition, while audit false-assurance rates remained non-zero. Among parser-valid degraded escalation outputs, recovery was 0.0 in every degraded condition. Detection of a degraded evidence state therefore did not imply restoration of successful operation under the implemented mechanism.

At the sequence level, 223 of 240 condition-linked groups met the operational cascade-failure definition. The corresponding taxonomy separated parser failure, detected-but-unrecovered degradation, audit false assurance, incomplete execution, and preserved success. This sequence-level view complements the paired intervention analysis by distinguishing failure families that imply different technical responses.

Taken together, the findings support the central argument of the paper: parser validity is necessary for automated pipeline operation but is insufficient as a proxy for Evidence-State Reliability. Evaluating whether an output satisfies a structural contract does not establish that the evidence available to support the substantive function of a pipeline stage remains adequate.

The contribution is therefore a bounded methodological and empirical operationalisation of controlled evidence-state intervention, stage-aware reliability measurement, parser-versus-evidence divergence, separation of audit detection from recovery, matched uncertainty analysis, and sequence-level reliability diagnosis.

These conclusions remain limited to one model configuration, one pipeline design, one scaled execution run, 60 selected sanitised cases, three controlled degradation families, and model-output-coded scoring. Cross-model, multi-run, multi-domain, and independently adjudicated replication is required before broader claims concerning generality, deployment reliability, or institutional assurance can be supported.

\bibliographystyle{plainnat}
\bibliography{references}

\end{document}